\documentclass{article}

\usepackage{microtype}
\usepackage{graphicx}
\usepackage{subfigure}
\usepackage{booktabs} 

\usepackage{hyperref}

\usepackage[accepted]{icml2023}

\usepackage{amsmath}
\usepackage{amssymb}
\usepackage{mathtools}
\usepackage{amsthm}

\usepackage[capitalize,noabbrev]{cleveref}

\theoremstyle{plain}

\theoremstyle{definition}

\theoremstyle{remark}

\usepackage[textsize=tiny]{todonotes}

\icmltitlerunning{Learning to acquire novel cognitive tasks}

\begin{document}

\twocolumn[
\icmltitle{Learning to acquire novel cognitive tasks with evolution, plasticity and meta-meta-learning}



\icmlsetsymbol{equal}{*}

\begin{icmlauthorlist}
\icmlauthor{Thomas Miconi}{yyy}
\end{icmlauthorlist}

\icmlaffiliation{yyy}{ML Collective}

\icmlcorrespondingauthor{Thomas Miconi}{thomas.miconi@gmail.com}

\icmlkeywords{Meta-learning, evolution,  neuromodulation, plasticity}

\vskip 0.3in
]



\printAffiliationsAndNotice{}  

\begin{abstract}

A hallmark of intelligence is the ability to autonomously learn new flexible, cognitive behaviors - that is, behaviors where the appropriate action depends not just on immediate stimuli (as in simple reflexive stimulus-response associations), but on contextual information that must be adequately acquired, stored and processed. While many meta-learning algorithms can design agents that autonomously learn new tasks, cognitive tasks adds another level of learning and memory to typical ``learning-to-learn'' problems. 
Here we evolve neural networks, endowed with plastic connections and neuromodulation, over a sizable set of simple cognitive tasks adapted from a computational neuroscience framework. The resulting evolved networks can automatically modify their own connectivity to acquire a novel simple cognitive task, never seen during evolution, from stimuli and rewards alone, through the spontaneous operation of their evolved neural organization and plasticity system. Our results emphasize the importance of carefully considering the multiple learning loops involved in the emergence of intelligent behavior.

\end{abstract}

\section{Introduction}

An important feature of intelligent behavior is the ability to learn not just simple, reactive tasks (associating a stimulus with a response), but also more complex, \emph{cognitive} tasks. While ``cognition'' in general is difficult to define precisely, here we operationally define ``cognitive'' tasks as those that require storing and manipulating a certain unpredictable piece of information for each new instance of the task - that is, tasks that require working memory. 

Under this definition, acquiring a new cognitive task is necessarily a special case of ``learning to learn'', or \emph{meta-learning}. Meta-learning, that is, the acquisition of learning-dependent tasks, has been studied quantitatively in animals; examples include Harlow's seminal study of memory-guided choice among pairs of objects \citep{harlow1949formation}, or more recently, mice learning to perform memory-guided navigation in virtual reality mazes \citep{morcos2016history}.

Many meta-learning algorithms exist for artificial agents \citep{thrun98learning,schmidhuber1993reducing,hochreiter2001learning,finn2017model,wang2016learning,duan2016rl2,bengio1991learning,floreano2000evolutionary,ruppin2002evolutionary,soltoggio2008evolutionary,miconi2016backpropagation,miconi2018differentiable,kirsch2019improving,kirsch2022introducing}. These algorithms produce agents that can successfully perform new instances of one specific learning-dependent (meta-)task (bandits, maze-solving, foraging,  etc.) Here, however, we seek something different. Rather than use an external algorithm to train an agent for this or that meta-task, we would like to build an agent capable of \emph{automatically}
acquiring novel, cognitive (memory/learning-dependent) tasks, including tasks never seen before in the agent’s lifetime
(or during its initial design), through the operation of its own internal machinery, from stimuli and
rewards alone - much like animals in the experiments mentioned above.


We propose to use an evolutionary process to design a self-contained network, endowed with plastic connections and reward-modulated plasticity. We expect that this evolved agent, when exposed to many episodes of a novel cognitive task, will automatically refashion its own connectivity in order to adequately extract, store and manipulate task-relevant information during each new episode of the task. Importantly, we do not want the agent to merely be able to perform tasks from a  pre-defined fixed set; rather, we want an agent that can automatically acquire novel cognitive tasks, including tasks never encountered during evolution.



To provide a sizeable number of computationally tractable cognitive tasks, we  use the formalism of \citet{yang2019task}, which implements a large number of simple cognitive tasks from the animal neuroscience literature (memory-guided saccades, comparing two successive stimuli, etc.) in a common format. 
One advantage of these tasks is that  each episode consists of only one trial (much like in the Omniglot task, a common meta-learning benchmark).
 In this 
paper, due to limited computational resources, we restrict this framework to only use binary stimuli and responses (see Methods). While this simplification considerably restricts the amount of learning that occurs in the innermost loop (within each trial/episode), it preserves the fundamental aspect of meta-learning and cognitive tasks: something new must be learned, memorized and exploited within each episode/trial. The framework can of course be extended to include any task that can be formulated as a series of stimuli-delay-response(s) trials, including more general meta-learning tasks (see Discussion). Thus, this framework combines computational efficiency with biological relevance.

Because we operationally define cognitive tasks as those that require some within-episode learning and memory, evolving such a cognitive learner necessarily involves at least three nested learning/memory loops (evolution, task learning, and episodic learning). The multiplicity of nested learning loops in nature, and the potential usefulness of extending typical meta-learning with additional levels, have already been pointed out before \citep{wang2021meta,miconi2019backpropamine} -  see Appendix \ref{sec:extendedrelated} for related work. Our results demonstrate the feasibility of this approach; they also suggest that acknowledging the multiple levels of learning involved in an experiment can actually have great practical impact on performance (see Appendix \ref{sec:noplast} and Discussion). 

\section{Methods}

\label{sec:methods}

\subsection{Overview}

Here we first provide a brief summary of the experiment. A complete description is provided in the following subsections. See also Figure \ref{fig:schema2loops} for an overview of the entire process. All code is available online (see Appendix \ref{sec:github}).

We are evolving a plastic (that is, self-modifying), fully-connected recurrent network to be able to autonomously acquire novel simple cognitive tasks. The whole experiment is composed of three nested loop. The outermost loop is the Evolutionary loop, which loops over lifetimes; the middle loop, or Task loop, constitutes the lifetime of the agent and loops over trials/episodes of a given task; and the innermost loop (the Episode loop) loops over timesteps. 

During each trial/episode, the network observes two stimuli in succession, produces a response, and then receives a reward signal (which depends on the task, stimuli and response for this trial), and applies reward-modulated synaptic plasticity to its own connections. The synaptic plasticity rule is guided by the evolved, innate parameters of the network, namely, its innate (``birth'') weights $\mathbf{W}$ and plasticity coefficients $\mathbf{\Pi}$ (as detailed below). The network is exposed to 400 such trials of a given task, constituting the Task loop, during which it is expected that the network will learn the task through the operation of reward-modulated plasticity, as guided by its evolved innate weights and plasticity parameters. 

Notice that these two loops (Trial/Episode loop and Task loop) are homologous to the ``inner loop'' and ``outer loop'' of a typical meta-learning experiment  (though with a very simple, but still  learning/memory-dependent, inner loop). The difference is that, in typical meta-learning experiment, the network modification occurring in the outer loop is determined by a fixed, hand-designed external algorithm. Here however, network modification is determined by an evolved, self-contained plasticity process, guided by the innate structure of the network, which is optimized by evolution in an additional, outermost loop. Specifically, the process described above is iterated (with a randomly chosen task from the training set for each pass) over a third, Evolutionary loop, in which evolution optimizes the network's structural parameters (innate weights $\mathbf{W}$ and plasticity parameters $\mathbf{\Pi}$) in order to improve the network's task learning ability (as estimated by the network's performance over the last 100 trials of the Task loop).

At regular intervals, we test the network on a withheld ``test'' meta-task, never seen during evolution, to assess the network's performance on the overall objective: automatic acquisition of \emph{unseen} cognitive tasks. In all experiments in the main text, the test task is  ``Delayed Match to Sample'' (DMS: are the two successive stimuli identical?). Results with other test tasks are reported in the Appendix.

\begin{figure*}
    \centering
    \includegraphics[scale=.25]{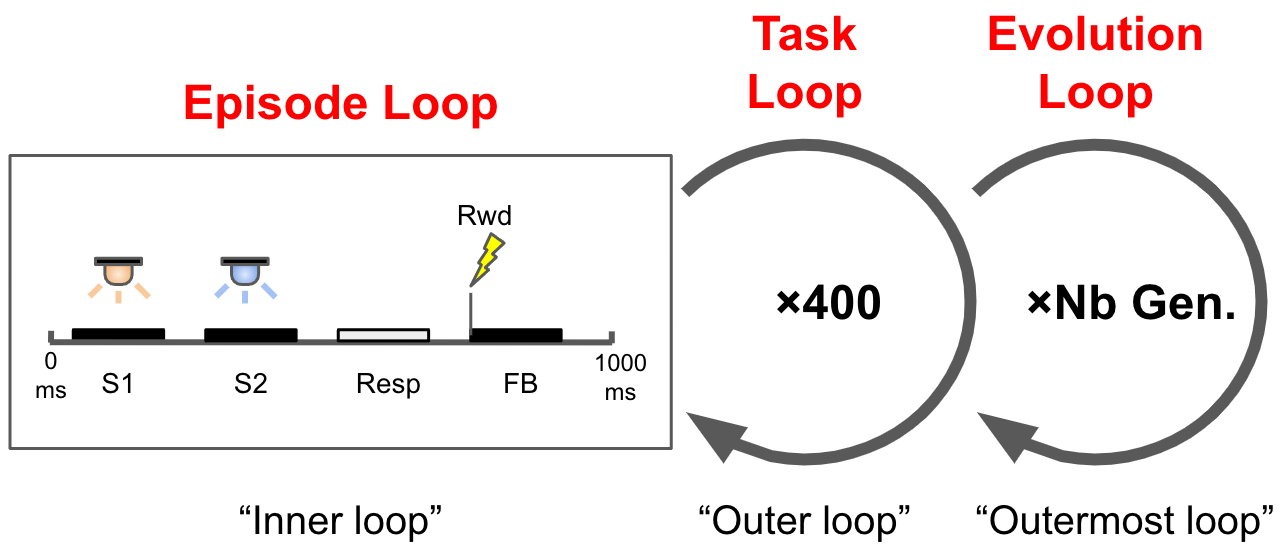}
    \caption{Overall organization of the experiment. During the Episode loop (which loops over time steps), the network stores and processes episodic information (stimuli S1 and S2) in its recurrent activations, provides a response (Resp) and receives a neuromodulatory reward and a feedback signal. In the Task loop, the network is modified by reward-modulated plasticity between episodes, storing task-solving information in plastic connections. These two loops correspond to the ``inner loop'' and ``outer loop'' of typical meta-learning experiments. An additional Evolutionary loop then optimizes the innate connectivity and plasticity structure of the network (that is, its genome), ensuring that the neuromodulated plasticity being applied after each trial results in proper lifetime learning.}
    \label{fig:schema2loops}
\end{figure*}


\subsection{Outermost loop: Evolution}

The evolved parameters (the ``genome'' of the agent) consist of a network's baseline (``innate'') weights $\mathbf{W}$ and plasticity parameters $\mathbf{\Pi}$, jointly denoted as $\theta = \{ \mathbf{W}, \mathbf{\Pi} \}$. $\mathbf{W}$ represents the initial weights at birth for the network, and $\mathbf{\Pi}$ represents the amount of plasticity at each connection.

The overall evolutionary algorithm is an evolution strategy similar to \citet{salimans2017evolution}. At each generation, the current candidate network $\theta_0$ is copied and randomly mutated many times, resulting in a population (batch) of mutated copies $\theta_i = \theta_0 + \sigma_i$ where each $\sigma_i$ is a Gaussian vector of the same size as $\theta$. Each of these mutated copies is then evaluated over a block of 400  trials, as described below, returning a loss $L_i$. Finally, the sum of all mutation vectors, weighted by their losses, is used as a gradient estimate $\hat{\nabla}_{\theta}L = \sum_{i} L_i \sigma_i$. As in \citet{salimans2017evolution}, we use antithetic sampling, which means that for every mutation vector $\sigma_i$, we also include the opposite mutation vector $-\sigma_i$ in the population (this considerably improves performance). We pass  this gradient  estimate  $\hat{\nabla}_{\theta}L$ to the Adam optimizer \citep{kingma2015adam} to produce actual parameter changes for $\theta$, resulting in a new $\theta_0$ for the next generation. We iterate this process for a fixed number of generations.

The reason we use evolution in the outermost loop, as opposed to some kind of backpropagation-based RL algorithm (as in \citet{wang2016learning} or \citet{duan2016rl2}) is simply due to the length of the lifetimes. As stated below, each  lifetime covers 400 trials of 50 time steps each, amounting to $20000$ total time steps. Backpropagating gradients over so many time steps seems unfeasible.

\subsection{Middle loop: Lifetimes/Tasks}

We evaluate each network by exposing it to a block of 400 trials of a given task. The task is chosen at random for each individual in the population, and the stimuli being presented during each trial are independently sampled for each individual at each trial (except that antithetic pairs receive the same tasks and sequence of stimuli). At the start of each block, a network is initialized by having its plastic weights (see below) and neural activities set to zero; the initial weights and plasticity parameters are taken from the mutated parameters $\theta_i$. Plastic weights, unlike innate weights, are updated over the whole block, without reinitialization, according to the plasticity rule described below. Neural activities and plastic weights are not reset between trials.

This loop corresponds to the \emph{outer loop} of a typical meta-learning experiment, during which the specific meta-task at hand (that is, the appropriate way to store and process task-relevant information during each inner-loop episode) is slowly learned across many episodes.

\subsection{Inner loop: Trials/Episodes}

\label{sec:trials}

During each trial, the network observes two stimuli in succession. These stimuli are fed to the network, one after the other (separated by a short delay),  by clamping a subset of neurons (the ``input'' neurons) to specific values for a fixed amount of time. Then, after another short delay, the network's response is recorded.  There are two output neurons, representing responses '0' and '1' respectively. The error for each trial/episode is computed as the absolute difference between the firing rates of the two output neurons (averaged over the response period) and their target for this trial.  

This error is then used as  negative reward to apply reward-modulated  plasticity at each connection ($-R$ in Equation \ref{eq:neuromod} below),  and also fed to the network as a stimulus, during the feedback period (FB in Figure \ref{fig:schema2loops}). Again, the error stimulus is provided by clamping the output of certain specific neurons (the ``feedback'' neurons) to the reward value for a fixed number of time steps. 

In addition, for analysis purposes, we  arbitrarily deem that a trial is ``correct'' depending on which of the two neurons has the highest mean firing rate over the response period (this is only for analysis and  visualization purpose and does not affect the algorithm itself in any way).

This process is iterated over 400 trials, constituting a block, or ``lifetime''. Then, the mean error over the last 100 trials of this lifetime block is used as the loss $L_i$ for each individual for this generation, guiding the evolutionary process as described above.

\subsection{Network operation and plasticity}

Plasticity in our model is based on the node-perturbation rule \citep{fiete2007model},  a reward-modulated Hebbian rule that models  the influence of dopamine on plasticity in the brain. Conceptually, node perturbation consists in occasionally and randomly perturbing the activation of neurons, then applying a reward-modulated Hebbian update to each synaptic weight, consisting of the product of the perturbation, the input at that synapse at the time of the perturbation, and the reward: $\Delta w_{i \rightarrow j}(t) \propto y_i(t) \Delta y_j(t) R$. The crucial element is that the ``output'' factor in this Hebbian rule is not the neuron's full output $y_j(t)$, but the (random) perturbation $\Delta y_j(t)$. 

 We  chose  this rule because it can be implemented in a reasonably biologically plausible manner \citep{miconi2016biologically}, yet comes with strong theoretical guarantees, since it is  largely equivalent to the REINFORCE algorithm \citep{williams1992simple} (the algorithm in section 5 of \citet{williams1992simple} is a node perturbation algorithm for stochastic spiking neurons). Furthermore, because it is a (reward-modulated) Hebbian rule, it fits easily with the differentiable-plasticity approach that we use in the evolutionary loop. 

Importantly, as we show in the following results, this plasticity rule in itself is insufficient to produce successful cognitive learning for the tasks used here. Rather, it serves as a basic building block that evolution uses to design efficient self-contained learners.

More precisely, at any time, the fully-connected recurrent network acts according to the following equations: 

\begin{align}
&\tau \frac{d\mathbf{x}(t)}{dt} = (\mathbf{W}+\mathbf{\Pi \odot P}(t))\mathbf{r}(t) - \mathbf{x}(t)\\
&\mathbf{r}(t) = f(\mathbf{x}(t))
\end{align}    

Here $\mathbf{x}$ is the vector of neural activations (the linear product of inputs by weights), $\mathbf{r}$ is the neural responses (activations passed through a nonlinearity), $\mathbf{W}$ and $\mathbf{\Pi}$ are the innate weights and plasticity parameters (fixed and unchanging during a lifetime and optimized across lifetimes by evolution, as mentioned above), $\mathbf{P}$ is the \emph{plastic} weights (changing over a lifetime according to the plasticity rule described below), $\tau$ is the network time constant, $f$ is a nonlinear function and $\odot$ represents the pointwise (Hadamard) product of two matrices. Note that these equations are just the standard continuous-time recurrent neural network equations, except that the total weights are the sum of innate weights and plastic weights multiplied by the plasticity parameters.

Random perturbations $\Delta \mathbf{x}(t)$ are occasionally applied to the activations $\mathbf{x}$. Furthermore, each connection maintains a so-called Hebbian eligibility trace $\mathbf{H}(t)$, which is a running decaying average of the product of modulations by inputs:

\begin{align}
   \tau_H \frac{d\mathbf{H}(t)}{dt} =  \Delta \mathbf{x}(t) \mathbf{r}(t)^\mathsf{T} - \mathbf{H}(t)
\label{eq:eligtrace}
\end{align}

Here  $\tau_H$ is the time constant of the eligibility trace, which is significantly longer than the neural time constant $\tau$. Finally, whenever a reward signal $R$ is applied (once per trial, after each response; see  section \ref{sec:trials}), the current value of the Hebbian  trace is multiplied by $R$ and added to the plastic weights $\mathbf{P}$:

\begin{align}
   \mathbf{P}(t) \leftarrow  \mathbf{P}(t) + \eta R \mathbf{H}(t)
   \label{eq:neuromod}
\end{align}

Here $\eta$ is the lifetime plasticity rate.  We reiterate that $\mathbf{P}(t)$ is initialized to 0 at the beginning of each block of trials (``lifetime'') and changes according to the above equations (without any reinitialization) during a whole block.

\subsection{Tasks}

In this 
paper, we simplify the framework of \citet{yang2019task} by only using binary stimuli and responses. The set of all possible tasks is the set of all 16 mappings from two successive binary stimuli to a binary response. Each trial consists of two binary stimuli shown in succession, each for a period of time; a response period, during which the network's response is recorded; and a feedback period during which the error signal is is provided to indicate whether the response was correct, and neuromodulated plasticity is applied. 

One task is withheld as the ``test'' task, which is never seen during evolution, and on which the evolved network is periodically tested to assess performance in the overall objective, that is, ability to acquire a \emph{novel} cognitive task. In the main text, our main withheld test task is the delayed-match-to-sample task (DMS, i.e. ``are the two successive stimuli identical or different?''). Results with different withheld test tasks are shown in the Appendix.  We chose the DMS task because it is the most difficult task to acquire in the set (see Appendix), and also because it is actually used in animal studies.  Note that the DMS task requires a strongly nonlinear processing of the two successive stimuli (being the negation of the exclusive OR). Thus, DMS is a simple, but not trivial task, with some degree of biological relevance. 

Importantly, we also remove the logical negation of the test task from the training set (for DMS, that is the ``delayed non-match to sample'' task), since the response for the logical negation of a task is simply the mirror image of the test task.   This is to ensure that the networks are truly able to learn the structure of the task at test time, rather than acquiring it through evolution and merely adapting the sign of the response at test time. This results in a training set of 14 different tasks for all experiments. 


\begin{figure*}[t]
    \centering
\includegraphics[scale=.66]{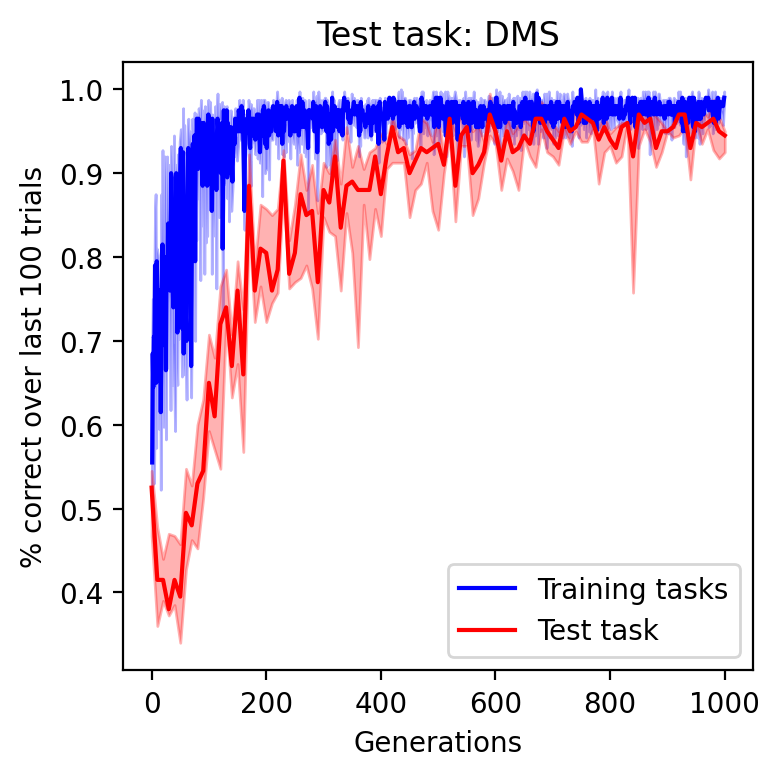}
\includegraphics[scale=.25]{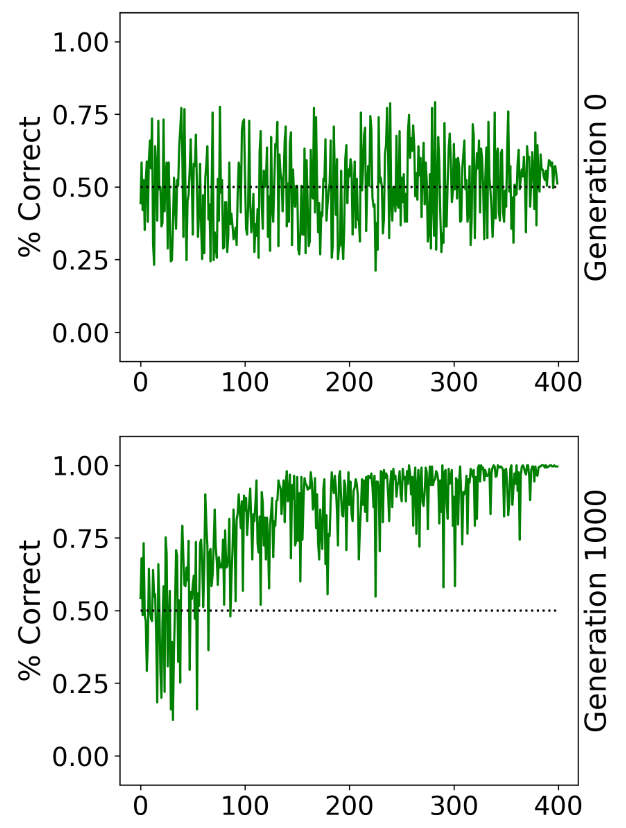}
    \caption{Left: Median (across all runs) proportion of correct trials, over the last 100 trials of each generation, shown separately for training tasks (blue, solid) and withheld task (red, dashed). Filled areas indicate inter-quartile range across runs. Right: Proportion (across a batch of 500 individuals) of correct trials over the course of a full block of the withheld task, shown separately at Generation 0 (i.e. randomly initialized network with uniform plasticity, top) and Generation 1000 (after evolving innate weights and plasticity of each connection, bottom). Dotted line: 50\% chance level. The evolved network successfully acquires the task over many trials/episodes. We stress that the 400 episodes shown here are equivalent to the \emph{outer loop} of a typical meta-learning experiment.}
    \label{fig:evoloss}
\end{figure*}

\subsection{Quantitative details and terminology}

Each generation is a batch of 500 individuals. Each block is composed of 400 trials, each of which lasts 1000 ms. Following \citet{yang2019task}, we use $\tau=100$ ms and simulation timesteps of 20 ms. Perturbations occur independently for each neuron with a probability of 0.1 at each timestep; perturbations are uniformly distributed in the $[-0.5, 0.5]$ range. We set $\tau_H = 1000$ ms, $\eta=0.03$. At generation 0, $\mathbf{W}$ is initialized with Gaussian weights with mean 0 and standard deviation $1.5/\sqrt{N}$, where $N=70$ is the number of neurons in the network (this distribution ensures self-sustaining spontaneous dynamics in the initialized network \citep{sompolinsky1988chaos}), while all values of $\mathbf{\Pi}$ are initialized to 0.5. Evolution runs over 1000 generations for DMS. We feed evolutionary gradients to the Adam optimizer, with a learning rate of 0.003. 

Regarding terminology, we use the words ``task'', ``meta-task'' and ``cognitive task'' interchangeably. An individual, or agent, means an individual network within the batch (``batch'' and ``population'' are also used  interchangeably). Due to the nature of our tasks, we also use ``trial'' and ``episode'' interchangeably: each trial/episode loop consists of the successive presentation of two stimuli, the network's response, and the feedback returned to the network, as shown in the ``Episode Loop'' section of Figure \ref{fig:schema2loops}. Each agent 's 
 lifetime thus consists of one block of 400 such trials/episodes of one randomly chosen task from the set (in the present design, each lifetime only contains one task).

\section{Results}

\subsection{Performance}

To assess overall progress, we track the evolutionary loss (that is, the total lifetime loss over the last 100 trials of each successive lifetime) as a function of the number of generations. Every 10 generations, we test the current candidate genotype $\theta_0$ on a withheld task, which does not result in any weight modification. We report both training loss (on the tasks used for evolutionary training) and test loss (on the withheld task), over 6 runs with different random seeds. 


As seen in Figure \ref{fig:evoloss}, the system successfully evolves an architecture that can automatically acquire a novel, unseen meta-learning task (left panel, red dashed curve). 

Importantly, initial performance at generation 0 (with random $\mathbf{W}$ and uniform $\mathbf{\Pi}$) is poor.
This confirms that the basic building blocks  of the system (plasticity, rewards and recurrence) do not suffice, by themselves, to produce a workable learning system. It is necessary to sculpt the structure  of the network (innate weights and plasticity coefficients), through an outermost loop, in order to turn these basic components into a successful self-contained learner (we confirm this with an additional experiment in Appendix \ref{sec:singlegen}). 

This can also be observed by tracking the lifetime performance (that is, the correctness of the response for each successive trial over a block) of the initial genotype at generation 0, which indicates little lifetime learning (Figure \ref{fig:evoloss}, right panel, top).

By contrast, after evolution has sculpted the innate parameters of the network, the candidate genotype now supports competent lifetime acquisition of the unseen meta-learning task. Again, this is confirmed by observing the lifetime performance of the evolved candidate, which indicates successful lifetime learning based on incoming correctness signals and synaptic plasticity (Figure \ref{fig:evoloss}, right panel, bottom).



\subsection{Evolutionary outcomes}

\label{sec:evooutcomes}

Importantly, the increase in performance is not simply due to a mere indiscriminate increase in overall plasticity. In a representative run, values in $\mathbf{\Pi}$, starting from a uniform value of 0.5, evolved to a final range of $0.0$ to $1.83$ (median $0.47$). Evolved values of $\mathbf{W}$ cover a larger range than their initialization values: in a representative run, weights went from an initial range of $[-0.71, 0.65]$ (median $-0.00$, median absolute value $0.11$) to a range of $[-1.60, 1.77]$ (median $0.00$, median absolute value $0.30$). The learned plastic weights $\mathbf{P}$ after lifetime learning (in a fully evolved individual) cover a smaller range than innate weights: at the end of the last block of a representative run (on the withheld task), values in $\mathbf{P}$ ranged from $-0.11$ to $0.11$ (median $0.00$). 

We did not detect any obvious structure in the evolved weights or plasticity parameters, or in the learned plastic weights (Figure \ref{fig:evolvedwandpi}). They do not seem to show specially different values for output or input neurons (whether on input or output connections), and the matrix of plasticity components shows no vertical, horizontal or diagonal structure. Evolved plasticity components look largely random besides the relatively constant overall level. 


\subsection{Importance of plasticity}

\label{sec:noplast}

\begin{figure}
    \centering
    \includegraphics[scale=.75]{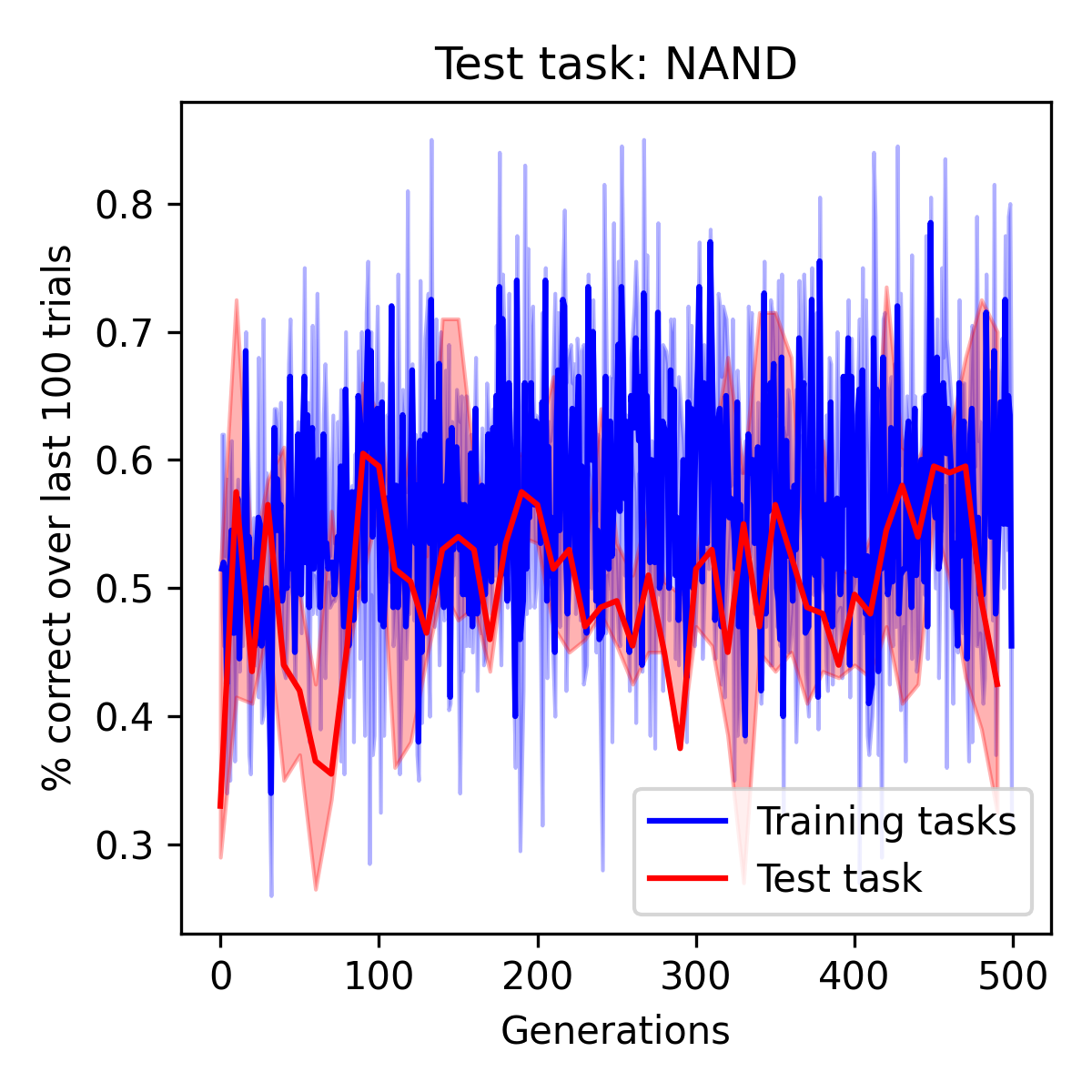}
    \caption{Same as Figure \ref{fig:evoloss} (Left), but with plasticity disabled. The network fails to learn either the training tasks or the withheld task, demonstrating the necessity of all three learning systems.}
    \label{fig:noplast}
\end{figure}

Because the network receives error signals, and because recurrent networks are universal approximators, it is theoretically possible that a non-plastic network might learn to solve the problems solely through its fixed weights and activities, as shown in typical meta-learning studies based on optimizing recurrent networks \citep{wang2016learning,duan2016rl2}. To test this possibility, we ran the exact same experiment,  but disabling plasticity (i.e. removing the $\mathbf{\Pi \odot P}(t)$ term in Equation 1). We also use an easier task as  withheld task, namely the  NAND task.  As expected given the small size of the network, the large number of tasks, and the multi-level nature of the experiment, this two-level system fails to acquire either the training tasks or the withheld task (Figure \ref{fig:noplast}). 



This confirms that all three learning/memory systems (evolution, plasticity and recurrence) are necessary for this model to succeed (the need for an outermost evolutionary loop is shown by the poor result of the unevolved, generation-0 networks in Figure \ref{fig:evoloss}). In turn, this shows the importance of considering all loops of learning that occur in a given experiments (see Appendix \ref{sec:noplastappendix}).

\subsection{Visualizing evolved and learned representations with cross-temporal decoding}

How does the evolved network perform its task? To investigate how evolved networks process information during each trial, we use cross-temporal decoding, a standard method in computational neuroscience \citep{meyers2008dynamic,king2014characterizing,stokes2013dynamic,miconi2016biologically}. For every pair of instants $t_1, t_2$ within a trial (1000 ms), we try to decode task-relevant information from neural activities at time $t_2$, based on a decoder trained on neural activities from time $t_1$. This allows us to determine whether the network encodes this information at any given time (by evaluating decoding performance when $t_1=t_2$), but also to estimate whether the network uses a stable encoding, or a dynamic, time-varying representation of task information: if the same decoder trained  on neural data from time $t_1$ can successfully decode task information from neural data at time $t_2$, this implies a similar encoding of this information at both time points (see Appendix \ref{sec:crosstemporal} for implementation details). 

We perform this analysis both for at generation 0 and generation 1000, as well as at the first and last trial of each generation. Note that this  allows us to separately asses the effect of evolution (by comparing generations 0 and 1000) 
 and of lifetime learning (by comparing the first and last block of each generation).
 
\begin{figure*}[t]
    \centering
    \includegraphics[scale=.66]{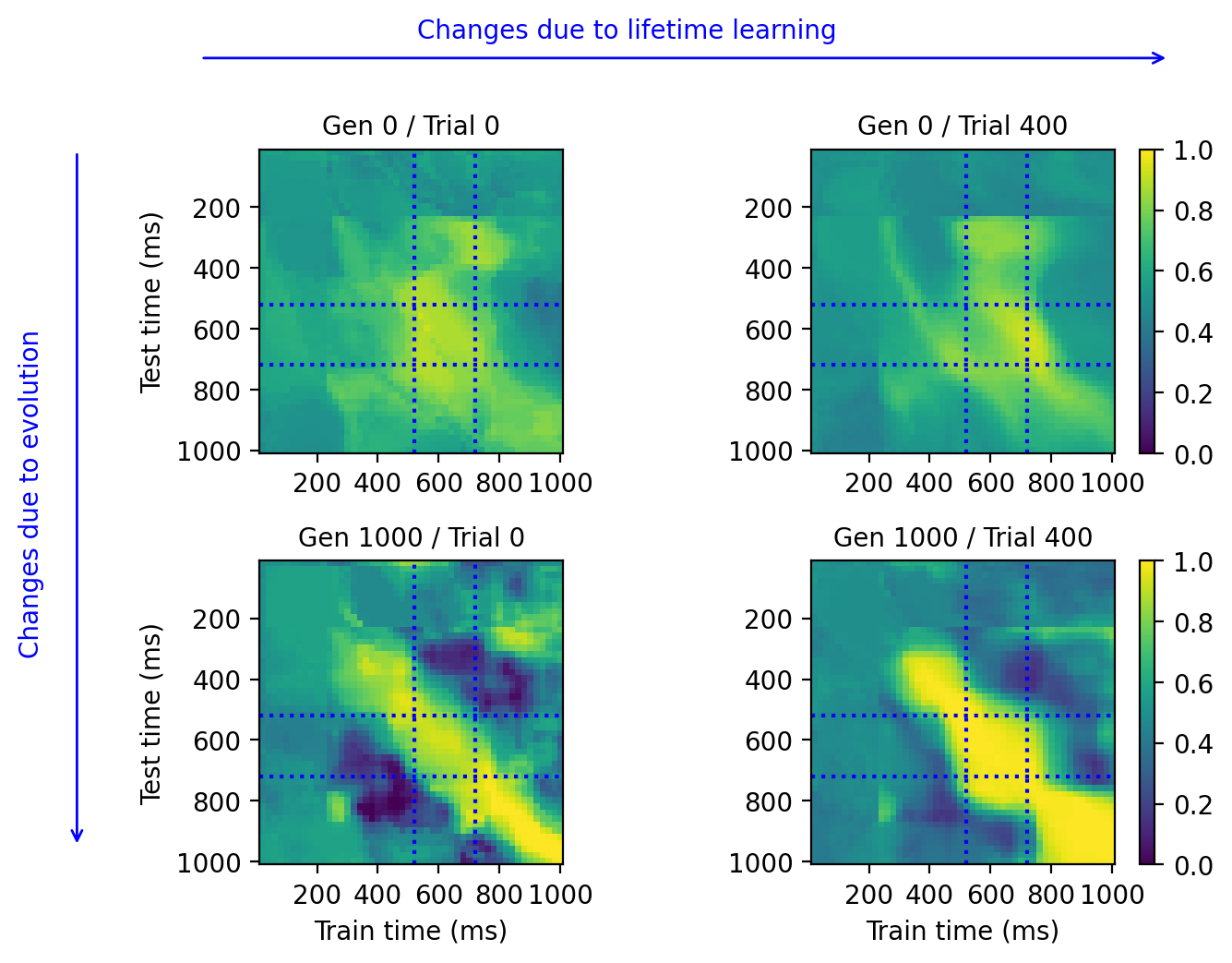}
    \caption{Cross-temporal decoding of target response for the (withheld) DMS task. Each matrix represents a full trial, at  generations 0 and 1000 (top  and bottom), and first and last trial of each generation (left and right). Dotted lines represent the response period, i.e. the 500-700ms interval. Each data point $x,y$ represents decoding performance at time $y$, using a decoder trained on neural  activities from time $x$. The network does not reliably encode target response (``are the stimuli equal?'') at generation 0, but does after evolution; furthermore, lifetime learning leads to a broadening of stable representations around response time (bottom-right plot). See also Appendix \ref{sec:crosstemporal} for similar plots for decoding identity of first and second stimulus.}
    \label{fig:decodingtarget}
\end{figure*}

 In Figure \ref{fig:decodingtarget}, we show decoding performance for target response (i.e. whether the two stimuli are identical or not), using  data from the same run described in Section \ref{sec:evooutcomes}. We see that generation-0 networks do not reliably encode this information at any time during the trial. By contrast, after evolution (generation 1000), the network robustly encodes this information, even though it was not part of its evolutionary training  (high decoding performance along the diagonal, $t_1=t_2$). Furthermore, comparing first-trial and last-trial data, we observe a stabilization of target response representation around the time of the  response period (dotted blue lines): the region of high performance bulges out into a squarish shape at this point. Interestingly, we observe quite different  dynamics when trying to decode stimulus identity rather than target response (see Appendix \ref{sec:crosstemporal}).

\section{Discussion}

\subsection{A multi-loop model for the emergence of autonomous cognitive learning}

In this work we sought to evolve a self-contained learning system, capable of acquiring novel simple cognitive tasks from mere exposure to stimuli and rewards. Because we operationally  define ``cognitive'' tasks as those that require learning a new piece of information for each new  trial/episode of the task, acquiring such a task is necessarily a special case of learning-to-learn (or at least ``learning-to-memorize'') problem. This adds another level of learning/memory in comparison to usual  meta-learning experiments.

The tasks used here represent a restricted form of meta-learning, such that each episode/trial is itself a learning item: something novel must be learned and exploited anew within each episode/trial, and the ability to perform this learning is optimized in the next-higher (``task'') loop. This is in contrast with usual meta-learning experiments: for example, when meta-learning maze-solving  \cite{duan2016rl2,miconi2018differentiable}, even when each individual task (i.e. each single maze) is run over several trials/episodes, the thing being learned over successive episode is always the same (this particular maze). The episodes in such tasks essentially serve as  ``macroscopic time steps'' in the task loop, rather than being  independent learning items of their own (see Appendix  \ref{sec:tableloops}). 

We acknowledge that, due to the extreme simplicity of the stimuli and responses, the individual tasks used here do not quite demonstrate \emph{general} meta-learning. Simple cognitive tasks with binary stimuli arguably constitute ``learning to memorize'' rather than truly ``learning to learn''. Nevertheless,  we believe this difference is largely quantitative: in structure, the tasks are mostly homologous to existing meta-learning problems (such as Omniglot and the Harlow task - see Appendix \ref{sec:equivalence}) and differ mostly by the simplicity and paucity of stimuli and responses per episode. Simple extensions (multi-dimensional stimuli, multiple responses per episode) would extend this framework to include common full-fledged meta-learning tasks. As such, we believe this work constitute an important - and necessary - step towards the evolution of agents capable of automatically acquiring novel full-on meta-learning tasks - that is, towards general meta-meta-learning.

Our model relies on multiple nested loops of learning, with each loop designing the learning algorithm of the next one, extending the standard two-loop model of meta-learning. The multiplicity of learning loops in nature has been observed before (see Appendix \ref{sec:extendedrelated}). 
Our results demonstrate the applicability of such a process to artificial agents, for the simple cognitive tasks described here. 

Importantly, our results show that the concept of multiple learning loops is not merely academic, but has practical importance: paying attention to the multiple levels of learning in an experiment can greatly impact performance, and be the difference between success and failure (compare Figure \ref{fig:evoloss} with Figure \ref{fig:noplast}; see Appendix \ref{sec:noplastappendix}). These results emphasize the importance of thinking carefully and systematically about the multiple loops and levels of learning, in nature or artificial experiments (see Appendix \ref{sec:tableloops}).

Although our initial results confirm the basic validity of the approach, it is clear that the present experiment is essentially a starting point. Building upon this platform, several avenues of investigation that may greatly enhance the abilities of the system immediately suggest themselves (see Appendix \ref{sec:extensions}). One important avenue of future research is the incorporation of an additional learning loop that is currently not included in the model, namely the  \emph{lifetime experience} loop: In the current form of the model, each lifetime is devoted to one single task. This is of course unrealistic. In reality, animals acquire a considerable amount of knowledge from their lifetime experience, accumulated across many different tasks, which greatly improves their performance in mature adulthood. This additional ``Lifetime'' loop would be crucial in studying the emergence of mechanisms that support robust \emph{continual learning} over a lifetime (forward and backward transfer, robustness against catastrophic forgetting, etc.). See Appendix \ref{sec:missingloop} for a discussion of this point.

\subsection{The ``Bitter Lesson'', AI-GAs, inateness, and the many loops of learning}

Why would one want to evolve a self-contained  cognitive learner, rather than simply apply an off-the-shelf meta-learning algorithm to any new cognitive task? First, for the same  reason that we use  meta-learning in the first place: the meta-meta-learning process may extract common structure in the environment and generic concepts useful across cognitive tasks, facilitating individual task learning \citep{wang2021meta}. 


More generally,  \citet{sutton2019bitter} has pointed out the ``bitter lesson'' that approaches leveraging learning and mass computation consistently overtake approaches based on hand-designed features and expert knowledge.  \citet{clune2019ai} pushes this argument further by proposing that the learning systems themselves (both their architecture and their algorithms) should be learned rather than designed, resulting in so-called ``AI-Generating Algorithms''. The present model applies these principles further by putting as much as possible  of the system  under the control of optimization rather than human design.  

In the other direction, \citet{marcus2019rebooting} have emphasized the importance of innate structure and ``common sense'' knowledge in the performance of human intelligence. The present model exemplifies that when the multiplicity of learning levels and loops is duly considered, the two aspects may become more complementary: one level's innateness is another level's learning, and one loop's structural knowledge is another loop's slowly acquired information (as pointed out by \citep{wang2021meta}). Furthermore, the present model affords considerable flexibility in balancing hand-designed structure with learning at multiple levels (node-perturbation vs. Hebbian, internal vs. external rewards, etc.)


In addition to these utilitarian concerns, we suggest that the emergence of an agent capable of autonomously acquiring novel simple \emph{cognitive} tasks, through its own internal machinery, is of interest in and by itself. This represents a step toward the goal of ``banishing the homunculus'' (to use the phrasing of \citet{hazy2006banishing}), that is, to  eliminate the need for an artificial, human-designed external ``executive'', and endow the system with truly autonomous learning. Obviously this  goal is still some way off: as mentioned above, the current system still includes many design restrictions. Nevertheless, we believe the success of this experiment demonstrates the potential of the approach.

Another point in \citet{sutton2019bitter} is the importance of ``methods that  continue to scale with increased computation''. We note that the experiments described above are strongly resource-limited, and the several obvious possible extensions mentioned in Appendix \ref{sec:extensions} are very much computation-dependent. Thus, the framework described here is likely to benefit greatly from increased computational resources.


\bibliography{biblio}
\bibliographystyle{icml2023}

\newpage
\appendix
\onecolumn

\renewcommand\thefigure{A\arabic{figure}}    
\setcounter{figure}{0}

\section{Extended related work}

\label{sec:extendedrelated}

\subsection{Two forms of meta-learning}

As mentioned in the introduction, many meta-learning algorithms exist \citep{thrun98learning,schmidhuber1993reducing,hochreiter2001learning,finn2017model,wang2016learning,duan2016rl2,bengio1991learning,floreano2000evolutionary,ruppin2002evolutionary,soltoggio2008evolutionary,miconi2016backpropagation,miconi2018differentiable,kirsch2019improving,kirsch2022introducing} Interestingly, many (but not all) of these algorithms fall within one of two broad categories, supported by very different interpretations:

\begin{enumerate} 

\item Algorithms where the inner-loop stores episodic information in the time-varying neural \emph{activities} of a recurrent network, while the outer loop   slowly modified in order to optimize within-episode learning \citep{hochreiter2001learning,wang2016learning,duan2016rl2}. A biological interpretation of this method is that the inner loop represents the within-episode self-sustaining activity of cerebral cortex, while the outer loop represents lifetime sculpting of neural connections by reward-modulated synaptic plasticity, as it occurs in the brain under the effect of neuromodulators such as dopamine (this interpretation is made explicit by \citet{wang2018prefrontal}). 

\item  Algorithms where the inner-loop stores episodic information in the synaptic \emph{connections} of the network, through some kind of parameterized synaptic plasticity algorithm, while the outer loop optimizes the innate structure of the network and/or the parameters of  synaptic plasticity\citep{schmidhuber1993reducing,bengio1991learning,floreano2000evolutionary,ruppin2002evolutionary,soltoggio2008evolutionary,miconi2016backpropagation,miconi2018differentiable}. In a biological interpretation, the inner loop represents lifetime learning by the brain's synaptic plasticity, while the outer loop represents the \emph{evolutionary} design of this built-in synaptic learning system over many lifetimes (even though the actual outer loop algorithm may not be evolutionary \citep{schmidhuber1993reducing,miconi2018differentiable}).

\end{enumerate}

Each of these methods can be applied to acquire one single meta-learning task, after which the agent can automatically perform new instances (episodes) of this same task. However, note that these two types of methods have a clear point of junction: the outer loop of the former uses the same substrate as the inner loop of the former (namely, synaptic plasticity). This suggests a combined overall method, in which an outermost ``evolutionary'' process would design a self-contained, reward-based plastic network over many lifetimes; in turn, this evolved plasticity machinery, guided by within-episode rewards, would automatically sculpt the network's own connections over many episodes of any given memory-guided task (including new such tasks, never seen before), and thus refashion the network's connectivity to be able to extract, store and manipulate task-relevant information from the environment during each episode. The overall result would be the evolution of an agent capable of automatically acquiring novel meta-learning tasks, which we have argued here covers an important type of ``cognitive'' tasks. This objective, mentioned in \cite{miconi2019backpropamine}, is the purpose of the present paper.

\subsection{Interactions of evolution and lifetime learning}

Evolving agents capable of lifelong learning has a long history \citep{ackley1991interactions,hinton1987learning,floreano2000evolutionary,soltoggio2008evolutionary,soltoggio2013solving} (see \citet{soltoggio2017born} for a review). The interaction of evolution and lifetime learning gives rise to complex dynamics. One example is the Baldwin effect \citep{baldwin1896new,hinton1987learning}, under which learning can guide evolution: beneficial features that are initially learned during each lifetime become increasingly incorporated into the genome, as natural selection favors individuals ``born closer'' to the eventual, beneficial phenotype\footnote{In the words of \citet{baldwin1896new}: ``This principle secures by survival certain lines of determinate phylogenetic [i.e. evolved] variation in the directions of the determinate ontogenetic [i.e. learned] adaptations of the earlier generation  \ldots  So there is continual phylogenetic progress in the directions set by ontogenetic adaptation \ldots This complete disposes of the Lamarkian [\emph{sic}] factor''.}. This enhances the reach of evolution, allowing it to produce phenotypes that it would have been unlikely to hit upon alone \citep{hinton1987learning}. It also makes lifetime learning faster  and  more reliable - but also more constrained and less flexible (i.e. ``canalized'' \citep{waddington1942canalization}).

\subsection{Multiple loops of learning}


Meta-learning, or ``learning to learn'', generally involves  two nested learning loops: an ``inner loop'' which stores and process relevant episodic information during an episode of the task, and an ``outer loop'' which optimizes inner-loop learning and processing over may episodes. Several authors have proposed extending this hierarchy. \citet{schmidhuber1993neural} (also in many later  works) suggested  that a recurrent network  capable of outputting its own weight changes could in principle ``change not only themselves
but also the way they change themselves, and the way they
change the way they change themselves, etc.''. \citet{miconi2019backpropamine} suggested that meta-learning by weight optimization of recurrent networks \citep{hochreiter1997long,wang2016learning,duan2016rl2} could be extended with an additional level that would design the weight-modifying algorithm itself (rather than using a standard RL algorithm), pointing out the evolutionary design of humans (who are capable of meta-learning) as a  real-life example of such``meta-meta-learning''. \citet{wang2021meta} explicitly describes three nested ``learning loops'' in natural learning (evolution, meta-tasks, and specific tasks), with higher loops each providing structure and priors to optimize learning in the lower ones.

Here we posit that such a meta-meta-learning process is the natural way to design an agent capable of autonomously learning novel cognitive tasks. Using the framework of \citep{yang2019task} as a  source of multiple computationally efficient simple meta-tasks with some biological relevance, we successfully demonstrate this process. Furthermore, we show that the concept of multiple learning loops is not merely academic, but has real practical importance, as failing to account for the multiplicity of learning levels prevents successful optimization (see Figure \ref{fig:noplast}, Section \ref{sec:noplast}  and Appendix \ref{sec:noplastappendix}).

\section{Results with different test tasks}

\begin{figure}[t]
    \centering
    \includegraphics[scale=.45]{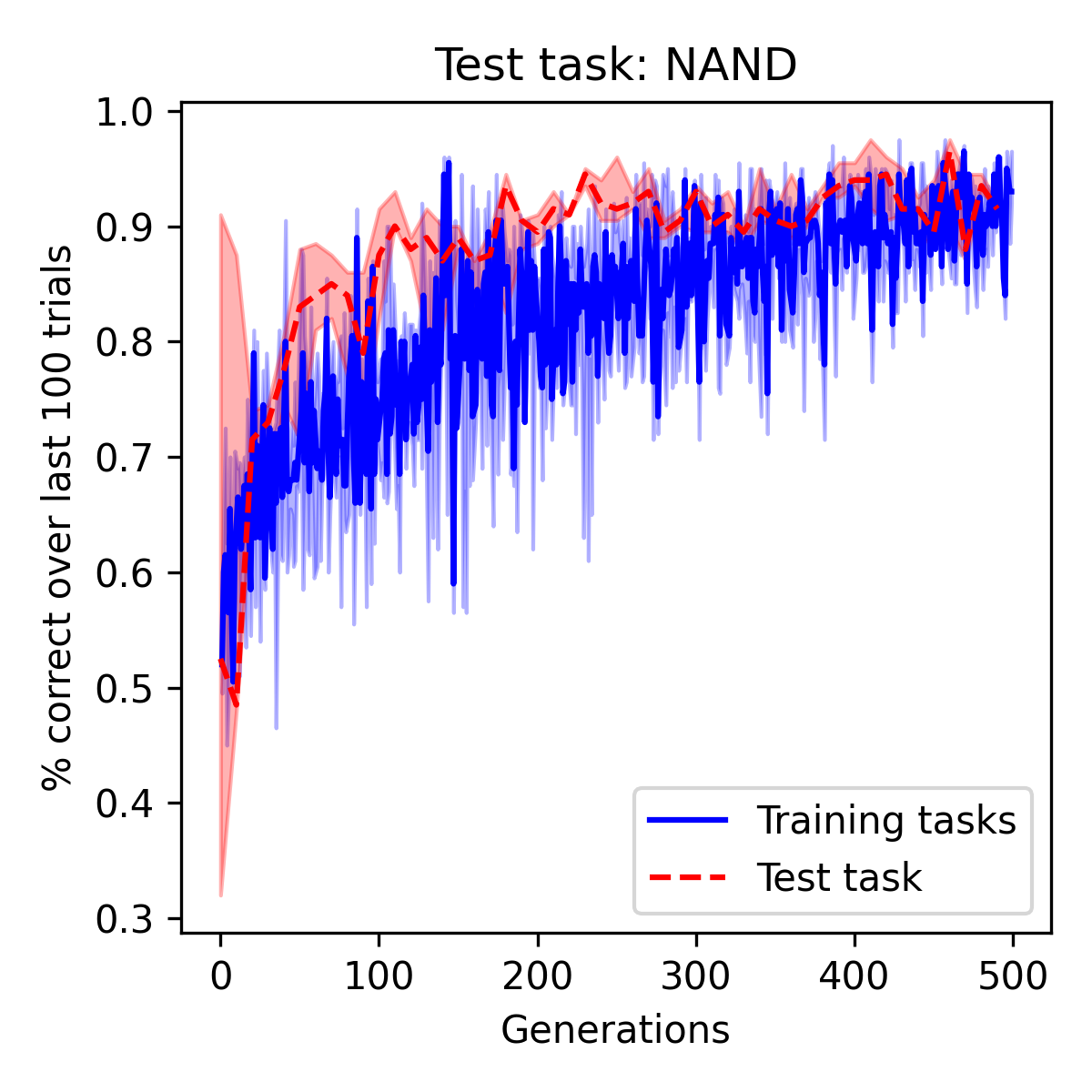}
    \includegraphics[scale=.45]{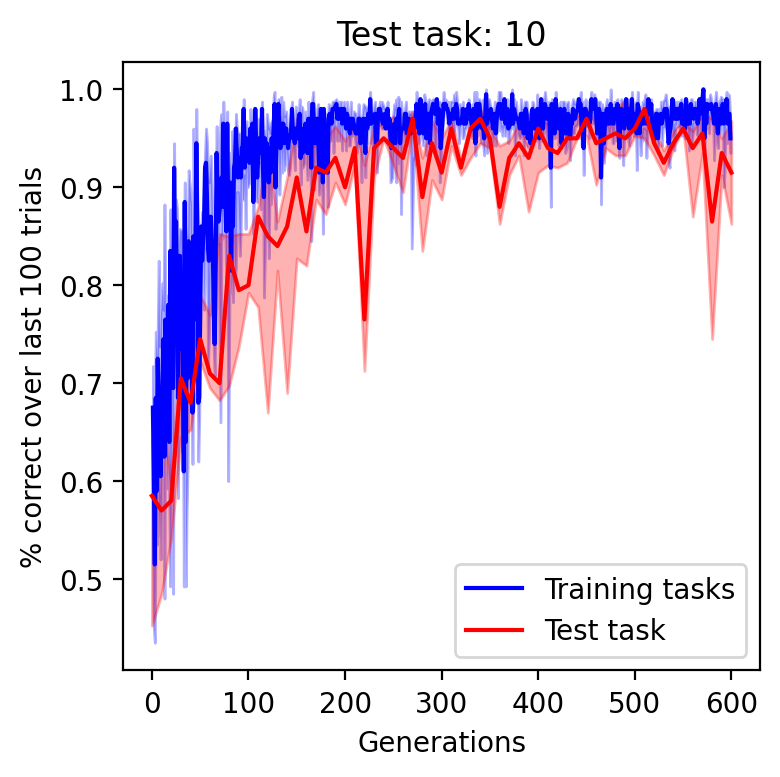}
    \includegraphics[scale=.45]{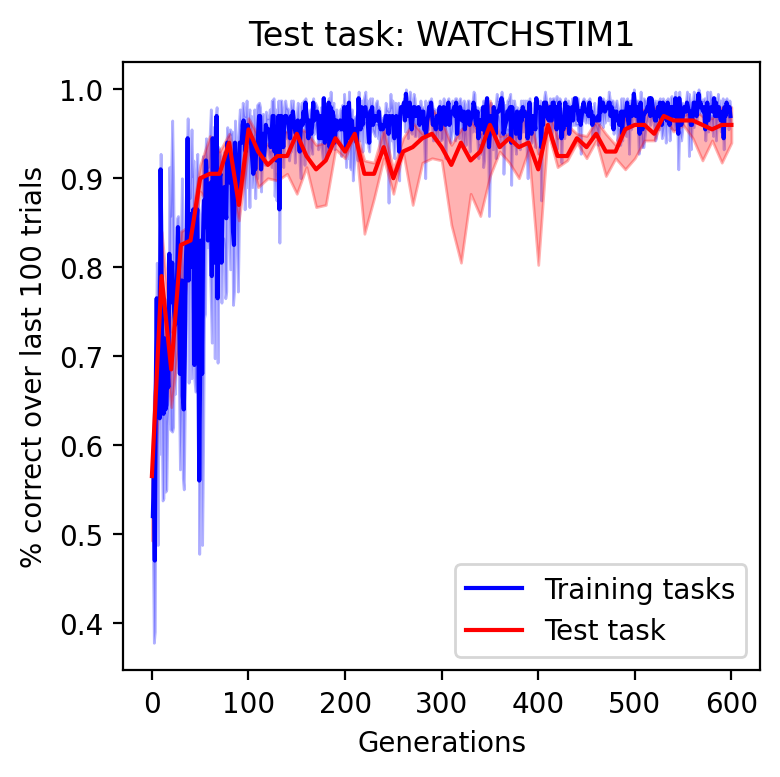}

    \caption{Performance over the course of evolution for different test tasks. Conventions are as in Figure \ref{fig:evoloss}, left panel. }
    \label{fig:evolossothertasks}
\end{figure}

In Figure \ref{fig:evolossothertasks}, we show results when using different tasks as the withheld test task. We report results with the NAND task, the ``report first stimulus'' task (respond with the value of the first stimulus, which requires memorizing this first stimulus in a way that is robust to the distractor second stimulus), and the ``1-0'' task (only respond positively if the two successive stimuli are  exactly 1 and 0). In all cases, as above, we also withhold the logical negation of the test task.

All of these tasks turn out to be much easier to acquire (as withheld test tasks) than the DMS task, with test performance occasionally higher than for the training set tasks (including DMS) that were used to guide evolution . This difference may reflect the inherent difficulty of the DMS task, which requires a non-linearly separable integration between the two successive stimuli.

\section{Results with disabled plasticity: the importance of multiple learning systems}

\label{sec:noplastappendix}

In Figure  \ref{fig:noplast}, we show the result of running the exact same experiment, but with disabled lifetime plasticity (removing the $\mathbf{\Pi \odot P}(t)$ term in Equation 1). In addition, we use an easier-to-acquire test task (logical NAND between the two successive stimuli). 

\begin{figure}
    \centering
    \includegraphics[scale=.75]{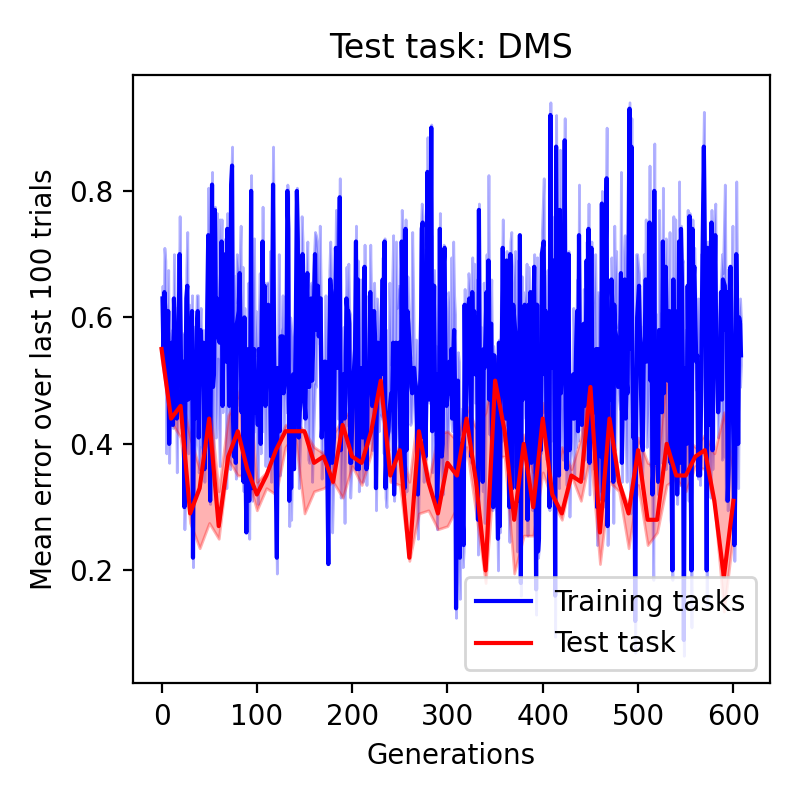}
    \caption{Results with plasticity disabled (as in Figure \ref{fig:noplast}), and doubling the number of neurons.  This larger model also fails.}
\label{fig:2xneur}
\end{figure}

In Figure \ref{fig:2xneur} we repeat this same experiment (removing plasticity), but  doubling the number of neurons (from 75  to 150), which doubles the number of parameters over the plastic network used in the main text (and also roughly doubles computation time  per generation). We see that the model fails similarly, showing that this failure is not a consequence of the small number of parameters.

Note that this control experiment only involves two levels of learning: evolution in the outer loop, and RNN operation in an undifferentiated inner loop over many  trials. As such, it is exactly equivalent to standard, two-loop meta-learning methods based on optimizing recurrent networks, such as RL$^{2}$ \citep{duan2016rl2} or Learning to Reinforcement Learn \citep{wang2016learning}.  We reiterate that we use evolution in  the outer loop instead of gradient descent, because it is not feasible to backpropagate gradients through twenty thousand time steps at each pass. 

As expected, this two-loop system fails to acquire either the training tasks or the withheld task (Figure \ref{fig:noplast}). This is in contrast with the success of our three-systems model, as shown in Figure \ref{fig:evoloss}.  This confirms that all three learning systems (evolution, plasticity and recurrence) are necessary for this model to succeed. 

Our hypothesis is that this negative result results from a mismatch between the two levels of learning in the system, and the three levels  of learning required by the experiment. Since our ``cognitive'' tasks require memorizing and manipulating information for each trial, learning an agent capable of acquiring such tasks necessarily involves three levels of learning/memory (see Figure A5).  By trying to shoehorn this process on only  two learning/memory levels (evolution/genome and RNN dynamics/activations), we are asking the recurrent networks to solve two levels  of learning by its operation: the within-episode storage and processing of episodic data, and the between-episodes learning of the actual task at hand through rewards. Evidently this is  beyond the capacities of such simple networks. By contrast, by acknowledging the three learning loops and introducing an additional learning system in the hierarchy (the plasticity system, which is optimized by evolution between lifetimes, and in turn optimizes the RNN between episodes),  the same networks now become capable of acquiring these tasks. This emphasizes the practical importance of considering the multiple loops of learning that occur in a given experiment \citep{wang2021meta}.

\begin{figure}
    \centering
    \includegraphics[scale=.75]{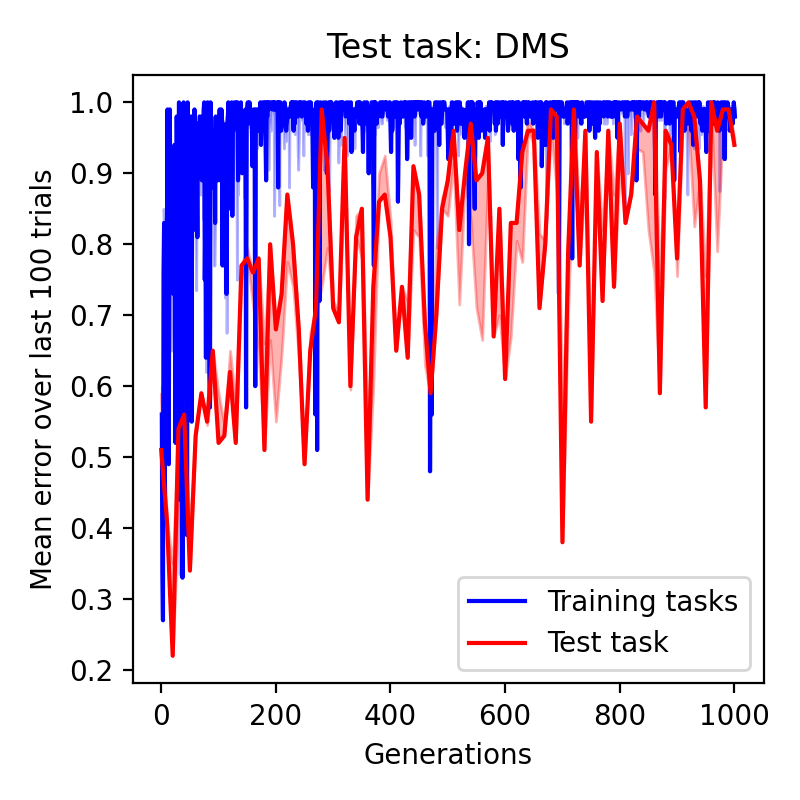}
    \caption{Results with plasticity disabled for all but 10 neurons (including the response neurons, but no input neurons). Although learning is less smooth than in Figure \ref{fig:evoloss}, the model still succeeds in developing task-learning  abilities.}
\label{fig:only10plast}
\end{figure}

Although plasticity only doubles the number of parameters and computation time, it offers the system an additional  $N^2$ degrees of memory (one per plastic connection) for storing ongoing episodic information. This is in contrast with non-plastic RNNs, which only have the  $N$ neurons to store such  within-episode information. To discern the role of this large potential memory capacity, we rand the same experiment as in the main text, but only allowing plasticity for 10 neurons, and only with each other (including the response neurons, but none of the sensory neurons). This was done by zeroing out  $\mathbf{\Pi}$ for all but the last 10 rows and 10 columns (the response neurons are always the last two neurons in numerical order). This results in $N+10^2  =  175$ degrees of freedom for storing real-time information, roughly in the same range as the 150 neurons of the large network in the previous section. We also found it useful to increase non-zero initial values of $\mathbf{\Pi}$ from 0.5 to 1.5. Remarkably, the model still manages to learn the training tasks and also develops the ability to acquire the withheld task, though less efficiently and with more noise (see Figure \ref{fig:only10plast}). This confirms that the benefit of plasticity does not merely result from a massive expansion  of memory storage (further experiments with removing plasticity for all neurons except between the two response neurons failed to perform above chance for the withheld task, as expected).

\section{Cross-temporal decoding}

\label{sec:crosstemporal}

Following an established method  in the neuroscience literature, we use a correlation-based classifier for decoding task-relevant information from neural activities \citep{meyers2008dynamic,stokes2013dynamic,king2014characterizing,miconi2016biologically} .  Each trial is performed in parallel by 250 networks. We divide this batch into 125 training runs and 125 testing runs. We separately average the network activities of the training set for each of the two possible target responses; for each time point, this gives us two average vectors of neural activities, each of which represents the ``prototype'' neural activity at this point for one of the two target responses. Then, for any pair of time points $t_1, t_2$ during the trial, we compute the correlation between the vector of neural activities at time  $t_2$ in each testing run with either of the two response-averaged training activity patterns (stereotypes) at time $t_1$, and pick the one with the  highest correlation as our decoded estimate of the target response for this testing run and this time point. We then compute the average accuracy of this decoding across all 125 test runs. This gives us the decoding performance value which is shown at point $t_1, t_2$ of the decoding matrices in the figure.

Note that correlation-based classifiers capture the information  that is contained in the collective pattern of the entire network, rather than isolated individual neurons, unlike (say) regression-based classifiers which might focus on specially informative individual neurons.

\begin{figure*}[t]
    \centering
    \includegraphics[scale=.4]{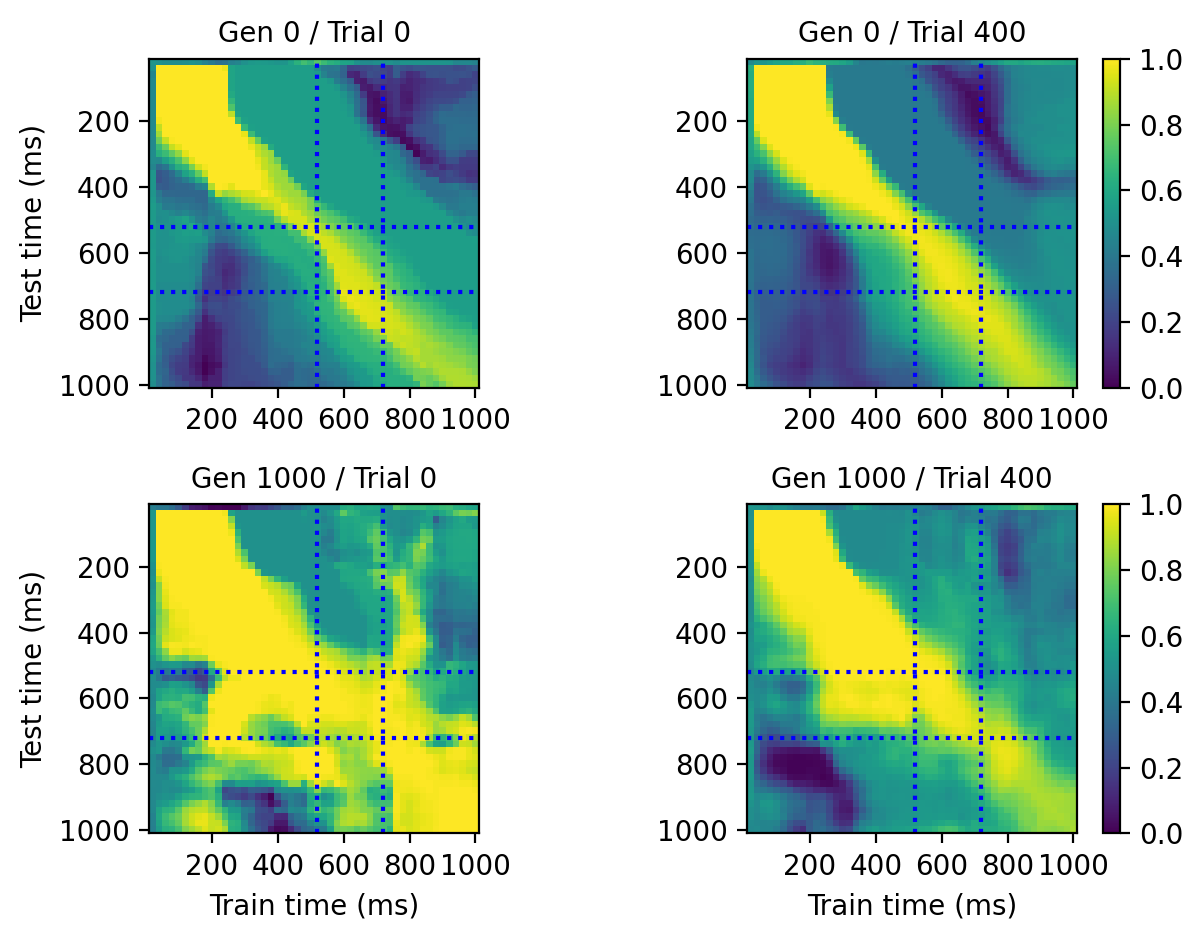}\hspace{1cm}
    \includegraphics[scale=.4]{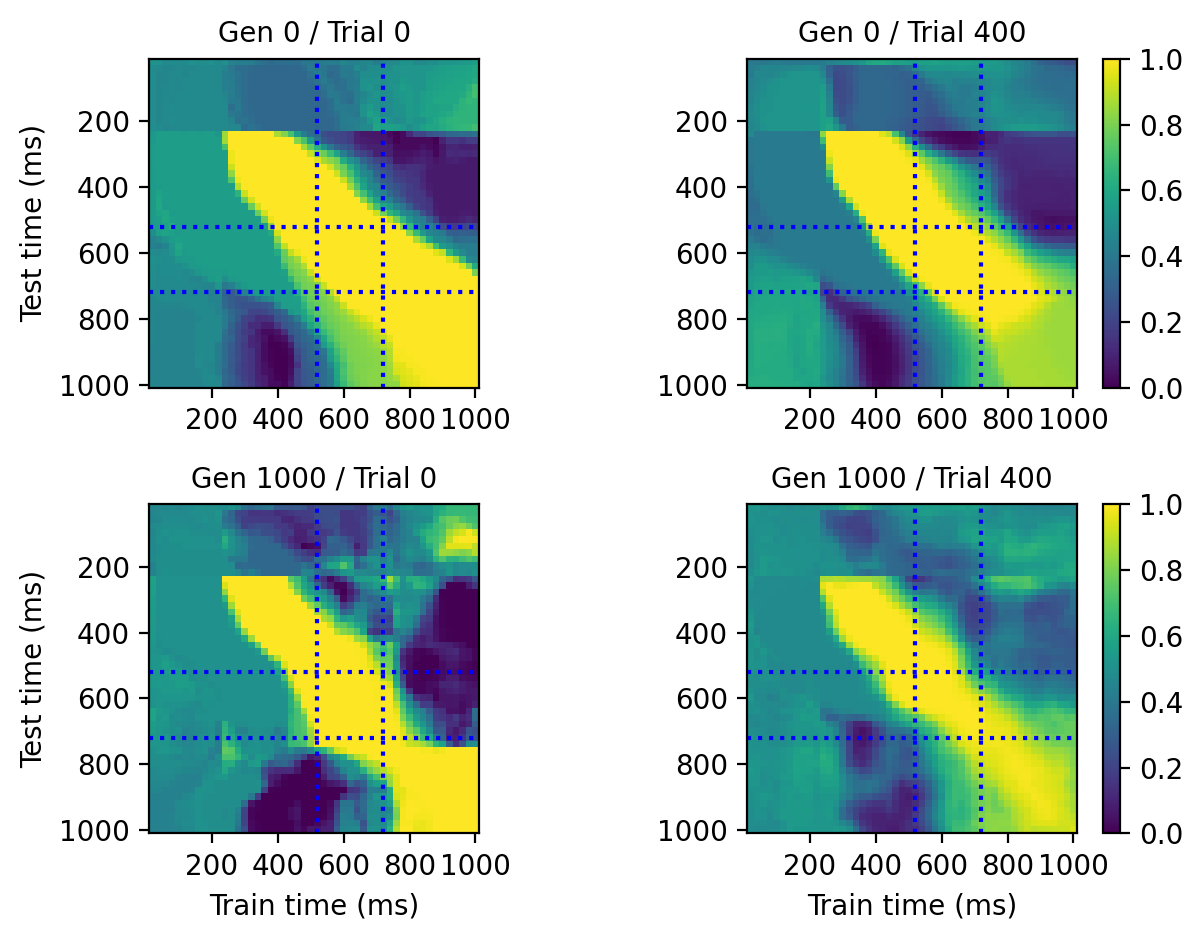}
    \caption{Cross-temporal decoding of first stimulus value (left) and second stimulus value (right) from neural activities. Conventions are as in Figure \ref{fig:decodingtarget}. See text for details.}
    \label{fig:decodingstim1stim2}
\end{figure*}

In Figure \ref{fig:decodingstim1stim2}, we show matrices of cross-temporal decoding for the identity of the first stimulus, and the identity of the second stimulus. We observe  a different pattern than for the target response in the DMS task  (Figure \ref{fig:decodingtarget}). First, even the generation-0, initialized network encodes this information to some extent, though the encoding is unstable and fleeting (top row). Furthermore, we see that after evolution, lifetime learning does not reinforce or stabilize encoding of stimulus identities during the response period; if anything, the encoding of stimulus identity during the response period seems somewhat less reliable after lifetime learning.  This is  what we would expect from task-appropriate learning, since the identity of each stimulus is irrelevant to producing the correct response (which depends on equality or inequality of the stimuli, independently of  their specific values).

\begin{figure}
    \centering
    \includegraphics[scale=.6]{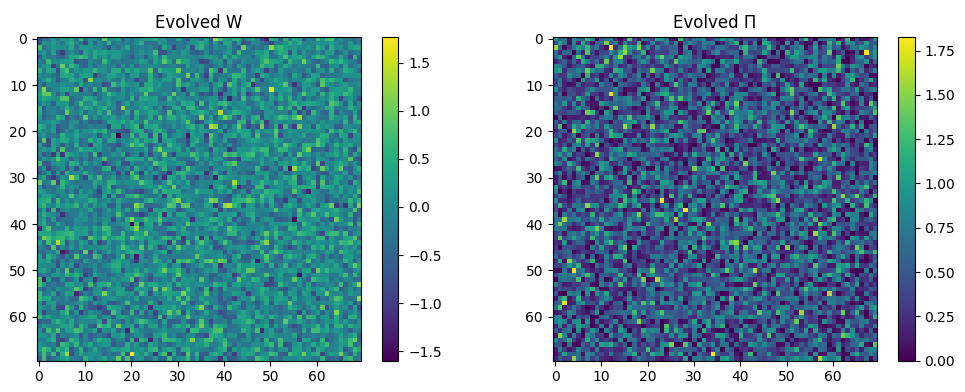}
    \caption{Final $\mathbf{W}$ and $\mathbf{\Pi}$ matrices (initial weights and plasticity coefficients of each connection) after evolution (data from the same run as in Figures \ref{fig:decodingtarget}, \ref{fig:decodingstim1stim2}).}
    \label{fig:evolvedwandpi}
\end{figure}

\section{Equivalence between the DMS task and a simplified Harlow task}
\label{sec:equivalence}

Because the tasks used here involve simple stimuli and operations, it is easy to overlook that they constitute actual meta-learning tasks, though of a restricted kind. To emphasize this point, here we show that the delayed-match-to-sample task (``same or different?'') is exactly identical to a simplified version of the Harlow task with two trials per episode.

In the seminal Harlow task \citep{harlow1949formation}, during each episode, a monkey must determine (by successive trial and error) which of two objects contains food. After enough episodes, monkeys are able to solve new episodes in just one trial, since the information received in the first trial immediately identifies which of the two objects contains the food.

Consider a version of this task where the same two objects are used for all episodes (which of the two objects is rewarded changes randomly for each episode), and each episode covers only two trials. Suppose also that the first choice is passive, that is, the object to be uncovered in the first trial is picked at random by the experimenter under the monkey's view, rather than by the monkey itself. In the second trial, the monkey chooses one of the two objects and gets to consume the food (if he chose correctly) as reward, as in the standard Harlow task. While these modifications make the task considerably simpler, they still preserve much of its overall structure - including the need to learn something unpredictable (which object is rewarded) for each new instance of the task. 

Yet this task is homologous our delayed match-to-sample task. The first binary stimulus represents which of the two objects is uncovered in the first trial. The second binary stimulus represents whether or not some food is present under that object. And the agent's response represents which object is to be selected at the second trial, receiving reward (food) if it chose the right object. As  the reader can verify, the correct response is exactly identical to that of the DMS task: to receive reward, the agent must produce response 1 if it saw stimuli 1 and then 1 or 0 and then 0, and response 0 otherwise. (Note that in this view, each episode represents two trials rather than just one.)

The point here is not to suggest that these simple tasks represent the full complexity of meta-learning. For example, the total set of possible sequences in each such task is strongly limited by the paucity and simplicity of stimuli, making it impossible to assess generalization.  Rather we wish to point out that they share an important element of general meta-learning, namely, the necessity to acquire and manipulate unpredictable information for each episode.  Furthermore, as mentioned in the Discussion, the framework can in principle be extended to more complex, unambiguous meta-learning tasks. The experiments discussed here, while restricted in scope, allow for a variety of simple cognitive tasks to be implemented with reasonable computational cost, while at the same time providing some biological relevance.

\section{Replacing evolution with within-lifetime plasticity}
\label{sec:singlegen}

In Figure \ref{fig:evoloss}, we see that performance at generation 0 (before evolution) is poor. This suggests that the basic building blocks provided to the system (recurrence, plasticity, rewards) do not suffice to produce an efficient learning system.

A possible objection is that the lifetime learning is much shorter than evolutionary time scale, and each lifetime only has access to  a single task. What if we extended a lifetime to cover as many successive tasks as the evolutionary process? Would we then see some overall successful learning, including the ability to acquire novel tasks, emerge from the raw building blocks? 

We performed the very same experiment, but removed evolution and instead used one very long lifetime, covering 1000 successive tasks. Importantly, we do \emph{not} reset the plastic weights between tasks, possibly  allowing learning to occur over successive tasks.  Regarding the test  task  (still DMS), we allow within-task plasticity to occur, but reset the plastic weights to the values they had immediately before the test task - that is, any learning occurring during the test task is immediately ``forgotten'' at the end of the test task. Thus, the test task is always ``novel'' to the system. Note that this would be equivalent to replacing meta-learning (in the two higher loops) with multi-task learning followed by fine tuning on novel tasks, an important subject of investigation \citep{mandi2022effectiveness}.

\begin{figure}
    \centering
    \includegraphics[scale=.75]{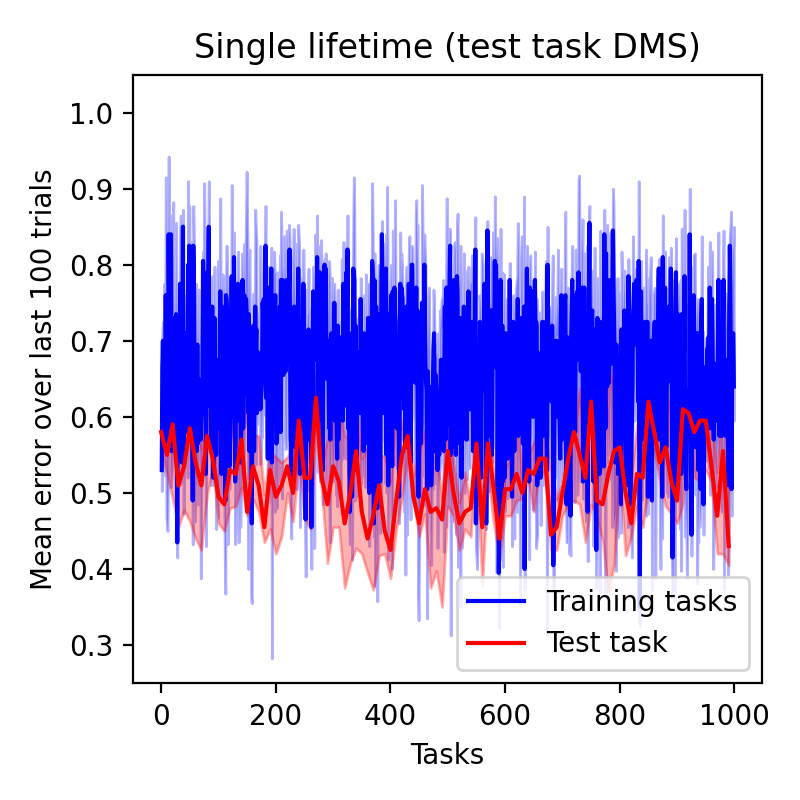}
    \caption{Replacing evolution with one single lifetime covering many successive tasks (plastic  weights are not reset between tasks). Medians and inter-quartile ranges over 6 runs are  shown both for training-set tasks  (blue) and test task  (red). The poor performance demonstrates the insufficiency of the basic building blocks and the necessity of evolution.}
    \label{fig:singlegen}
\end{figure}

However, as shown in Figure  \ref{fig:singlegen}, performance is poor, especially for the challenging test tasks (at chance).  This confirms that the basic components (recurrence, plasticity, reward signals), by themselves and without evolutionary sculpting, do not form a successful learning system.

\section{Extensions of the model}

\label{sec:extensions}

Here we describe ways in which the system can be extended to improve the resulting network's learning abilities.

\subsection{Tasks}

For computational efficiency reasons, the framework described above is a simplification of the one introduced by \citet{yang2019task}. In particular, we only use binary stimuli and response, as opposed to the real-valued (circular) stimuli and responses of the original study. \citet{yang2019task} also include many more tasks in their experiments, which have the added advantage of biological relevance (many of these tasks being classical tasks from the neuroscience literature). Thus, a straightforward way of improving our model is simply to make it more similar to the original framework of \citet{yang2019task}.

We emphasize that the tasks used here, despite their simplicity, are truly meta-learning tasks: new items of information must be acquired, stored and exploited appropriately at each episode. The fact that each trial is a full self-contained episode is similar to the Omniglot task (association between arbitrary stimuli and specific responses). In fact, a homologue of the Omniglot task itself can be implemented in this framework, if we extend the trials to include several, non-binary stimuli:  show two arbitrary ``cue'' stimuli; then the expected responses to each of these cues (arbitrarily chosen for each trial); then show one of the two cues; and finally record the network's response (which should be identical to the expected response for this particular cue, as shown previously). 

Furthermore, if we allow for more than one response (and return signal) per trial, we can also include reinforcement learning tasks, such as bandit tasks, in our task set, with each such ``trial'' still being a complete self-contained episode. Thus, despite its simplicity, the formalism can encompass all forms of meta-learning, including meta-supervised learning \citep{hochreiter2001learning} and meta-reinforcement learning \citep{wang2016learning,duan2016rl2}.

\subsection{Lifetime plasticity}

Perhaps the most simplistic aspect of our model is the handling of lifetime synaptic plasticity. While node perturbation is a powerful model of reward-modulated plasticity \citep{fiete2007model,miconi2016biologically}, it is obviously not meant to model the totality of synaptic plasticity processes; for example, it cannot implement pure Hebbian learning (under zero or constant rewards, the expected weight change is 0, which is indeed a requirement for reward-modulated Hebbian learning to perform successful reinforcement learning \citep{fremaux2010functional}). Furthermore, in the current system the modulatory signal $R$ is externally applied by the algorithm. A more realistic (and potentially more powerful) method would be to have the modulatory signal under control of the network itself, that is, making $R(t)$ an output of the network, as in \citep{miconi2019backpropamine}. 

In addition, in the model described above, the random perturbations of neural activity that support the node perturbation method are applied uniformly with a fixed probability. However, in nature, such sources of randomness are thought to be under control of dedicated brain structures, for example in models of bird  song learning \citep{olveczky2005vocal}. Putting randomness under control of the network is thus another possibility for future work.

\subsection{Non-biological alternatives}

Our model makes use of biologically inspired methods throughout (evolution, neuromodulated synaptic plasticity, recurrent neural networks,  etc.). We chose biologically-based methods both because of intrinsic interest, and because they allow wide flexibility in parametrizing the various systems involved, placing as much as possible under the control of optimization and learning. However, non-biological methods can also be incorporated into this framework. 

As an example,  a possible alternative would be to use backpropagation and gradient descent as the lifetime learning rule, in replacement of synaptic plasticity (assuming a supervised learning signal is available). In this setting, evolution would  guide the initialization of the network, while backpropagation through time would occur after each trial to optimize within-trial performance for the task at hand. We note that such a setting would essentially make the two upper loops identical to Evolutionary MAML (as in e.g. \citet{song2019maml}), with the bottom two loops being identical to L2RL / RL$^2$ \citep{wang2016learning,duan2016rl2}. Such a choice would trade off biological relevance and flexibility again potentially higher performance, at least for some supervised tasks.

Separately, while here we only consider memory mechanisms based on synaptic plasticity and recurrent activations, there are other, non-biological forms of long-term lifetime memory. In particular, information can be stored in explicit banks of embedded key-value pairs, accessed through some attentional mechanism, as in Neural Turing Machines \citep{graves14neural} and the MERLIN architecture \citep{wayne2018unsupervised}. The framework described here can be applied to such explicit forms of memory, in replacement of, or in conjunction with, synaptic plasticity, which might allow experimenters to assess the relative strengths of both approaches for the purpose of autonomous cognitive learning.

\subsection{The missing loop: lifetime experience}

\label{sec:missingloop}

\begin{figure*}
    \centering
    \includegraphics[scale=.3]{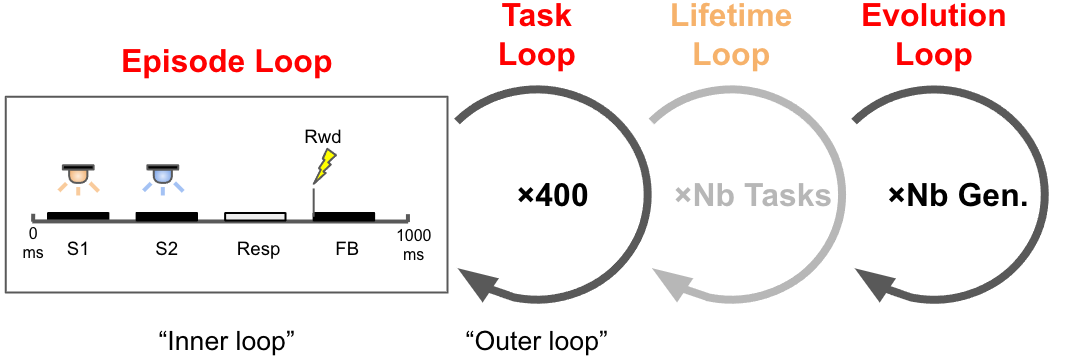}
    \caption{Overall organization of the experiment, with the (unimplemented) ``Lifetime loop'' added.}
    \label{fig:schema3loops}
\end{figure*}

\begin{figure*}[t]
    \centering
    \includegraphics[scale=.3]{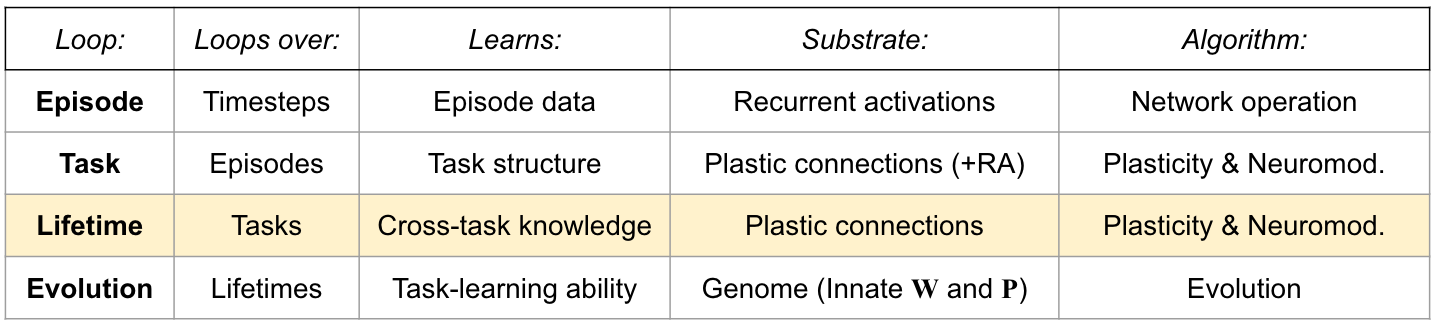}
    \caption{A classification table of learning loops, including their attributes.}
    \label{fig:tableloops}
\end{figure*}

In the current form of the model, each lifetime is composed of one block of trials. Each individual is initialized, performs one task, then is reinitialized again. This is of course unrealistic. In reality, animals acquire a considerable amount of knowledge from their lifetime experience, accumulated across many different problems, which greatly improves their performance in mature adulthood. In animal literature, examples of early life learning include the tuning of sensory cortex, the learning of songs in some bird species, etc. This long-term acquisition and transfer of lifetime experience is particularly developed in humans, in which it forms an important component of so-called ``common sense'' knowledge, that is thought to be critical to intelligent human behavior \citep{marcus2019rebooting}. 

In other words, the model described here deliberately excludes an additional learning loop, separate from the three loops considered here: the lifetime experience loop, set between the evolutionary loop and the task loop (see Figure \ref{fig:schema3loops}). This choice results from a desire for simplicity (and computational efficiency).
However, the present model can implement this additional loop by not reinitializing the networks between tasks, and adding more tasks in each lifetime. This may allow us to study the emergence of mechanisms that support robust \emph{continual learning} over a lifetime, including both forward transfer of information across tasks within a lifetime (``life experience'', including ``common sense'' - by adding more tasks per lifetime without reinitialization), and prevention of interference between learned tasks (``catastrophic forgetting'' - by repeating previously encountered tasks within each lifetime).

\section{Learning loops: a fundamental component in the emergence of intelligent behavior?}

\label{sec:tableloops}

One objective of this paper is to emphasize that multiple learning loops, nested into each other in a hierarchy  where each loop optimizes the learning of the previous one, govern the emergence of intelligent agents. This hierarchy extends the traditional dichotomy between ``inner loop'' and ``outer loop'' that is commonly observed in meta-learning,  as  suggested previously \citep{miconi2019backpropamine,wang2021meta}. 

As with standard meta-learning, thinking in terms of loops, and asking what is learned in each loop, how is it acquired, and how it is stored, can be helpful in managing the complexities of the process. 

One potential concern is the apparent risk of arbitrariness: whether a process is implemented as a loop is often an arbitrary decision by the designed of the experiment. For example, in a standard bandit meta-task, the meta-learning loops over episodes, while the episodes loop over trials (i.e. arm pulls). But what if the programmer decides to implement trials as temporally extended processes (moving towards the arm, grasping the arm, pulling it, etc.), looping over time steps? Does this arbitrary implementation decision add another learning loop to the process, and another ``meta-'' to the learning?

We found the following definitions useful:

\begin{itemize}
    \item A loop constitutes an actual learning loop if during each pass of the loop, some new, unpredictable piece of information is acquired, stored and exploited \emph{within the same pass} of the loop.
    \item Two loops are in a hierarchy (one is the ``meta'' of the other) if what is learned within one loop is expected to improve the learning in the other loop.
\end{itemize}

Note that these definitions address the example mentioned above: while the arm pulls may be implemented as ``loops'', what is learned during each such pull (an instantaneous return from one arm) can only be exploited in \emph{future} trials, or equivalently, in the same pass of the upper-level loop (i.e. the episode loop). Thus, in terms of learning, trials, even though they are implemented as loops, act as mere ``macro-timesteps'' of their upper-level loop (the episode loop), do not constitute an actual learning loop, and do not add a ``meta-'' to the process.

The fact that actual learning loops can be defined in a principled way, making them independent of arbitrary implementation choices, suggests that these loops are not mere incidental appearances, but constitute genuine, fundamental, objectively quantifiable components in the emergence of intelligent behavior.

Based on this framework, we can characterize each loop by asking specific questions about it, including:

\begin{itemize}
    \item What is being looped over?
\item What is being learned?
\item What is the substrate (that is, where is the thing to be learned stored)?
\item What is the algorithm?
\end{itemize}

We found these questions useful in preventing confusion between the multiple loops involved in natural learning. An example of this classification is shown in Figure \ref{fig:tableloops}.

\section{Code availability}
\label{sec:github}

All code is available at \texttt{https://github.com/ThomasMiconi/LearningToLearnCogTasks}



\end{document}